\documentclass[letterpaper, 10 pt, conference]{ieeeconf}  

\IEEEoverridecommandlockouts                              

\usepackage{amsmath} 
\usepackage{amssymb}  

\usepackage{csquotes}             
\usepackage{graphicx}             
\usepackage{float}                
\usepackage{bm}                   
\usepackage{color}                
\usepackage[table,xcdraw]{xcolor} 
\usepackage{multirow}             
\usepackage{hhline}               
\usepackage{pifont}               
\usepackage{makecell}             
\usepackage{tablefootnote}        
\usepackage{footmisc}
\usepackage{hyperref}             
\usepackage{url}                  
\hypersetup{
    colorlinks=true,
    linkcolor=blue,
    filecolor=cyan,      
    urlcolor=magenta,
    }

\definecolor{red}{RGB}{255, 0, 0}

\title{\LARGE \bf
Towards Human Haptic Gesture Interpretation for Robotic Systems
}

\author{Bibit Bianchini, Prateek Verma, and J. Kenneth Salisbury
\thanks{Bibit Bianchini is with the Department of Mechanical Engineering and Applied Mechanics, University of Pennsylvania, Philadelphia, PA 19104, {\tt\small bibit@seas.upenn.edu}}%
\thanks{Prateek Verma is an affiliate with the Artificial Intelligence Laboratory at the Department of Computer Science, Stanford University, Stanford, CA 94305,  {\tt\small prateekv@stanford.edu}}%
\thanks{J. Kenneth Salisbury is with the departments of Computer Science and Surgery, Stanford University, Stanford, CA 94305, {\tt\small jks@robotics.stanford.edu}}%
}

\begin{document}

\maketitle
\thispagestyle{empty}  
\pagestyle{empty}      

\begin{abstract}
    Physical human-robot interactions (pHRI) are less efficient and communicative than human-human interactions, and a key reason is a lack of informative sense of touch in robotic systems.  Interpreting human touch gestures is a nuanced, challenging task with extreme gaps between human and robot capability.  Among prior works that demonstrate human touch recognition capability, differences in sensors, gesture classes, feature sets, and classification algorithms yield a conglomerate of non-transferable results and a glaring lack of a standard.  To address this gap, this work presents 1) four proposed touch gesture classes that cover an important subset of the gesture characteristics identified in the literature, 2) the collection of an extensive force dataset on a common pHRI robotic arm with only its internal wrist force-torque sensor, and 3) an exhaustive performance comparison of combinations of feature sets and classification algorithms on this dataset.  We demonstrate high classification accuracies among our proposed gesture definitions on a test set, emphasizing that neural network classifiers on the raw data outperform other combinations of feature sets and algorithms.  Accompanying video is here.\footnote{Video: \url{https://youtu.be/gJPVImNKU68}}
\end{abstract}

\section{Introduction}
\subsection{Motivation and Applications}
An ultimate goal of the robotics sector, developing human-robot interactions (HRI) as natural as human-human interactions is most severely limited by physical HRI (pHRI).  While computer vision and natural language processing are strong research thrusts that provide eyes and ears for HRI, a robotic sense of touch faces the hurdle of no standardized parallel to cameras and microphones:  clearly established, off-the-shelf, and sufficiently reliable sensors.  The simple human task of identifying the meaning behind a human touch is a critical informative step to enable pHRI collaboration, as well as a significant challenge for robots.  Discerning among such subtleties with a contact sensor as ubiquitous as cameras and microphones are for capturing visual and auditory data, respectively, has great implications for the future of pHRI.


\subsection{Prior Work} \label{sub_sec:prior_work}
There are many works in the existing literature that demonstrate progress toward natural physical robot interactions in task-specific examples.  Using contact sensors, robots have been able to dance with their human partners \cite{chen2015evaluation}, perform handovers with humans \cite{Nagata1998, Konstantinova2017, he2015}, open a variety of doors \cite{jain2013improving}, insert pegs of unknown shape into precision holes \cite{li2019connecting}, and determine acceptable grip forces for grasping various objects \cite{Sadigh2009}.  Many physical robotic tasks are better informed by studying how humans accomplish them, such as for handovers \cite{Chan2012, Strabala2013, Mason2005, pan2017automated}.  However, the success of these demonstrations can be misleading, given the narrow capabilities of the robotic systems in such examples.  For instance, some HRI handover demonstrations require the human to apply aggressive forces for the robot to detect when to release a transferred object \cite{Nagata1998}, and others implement thresholding of single parameters, such as jerk \cite{Konstantinova2017}.  Even carefully tuned thresholds leave systems ample room to misclassify gestures, thus rendering the robotic system useless outside a narrowly-defined task.


\begin{figure}[t]
    \centering
    \includegraphics[width=\linewidth]{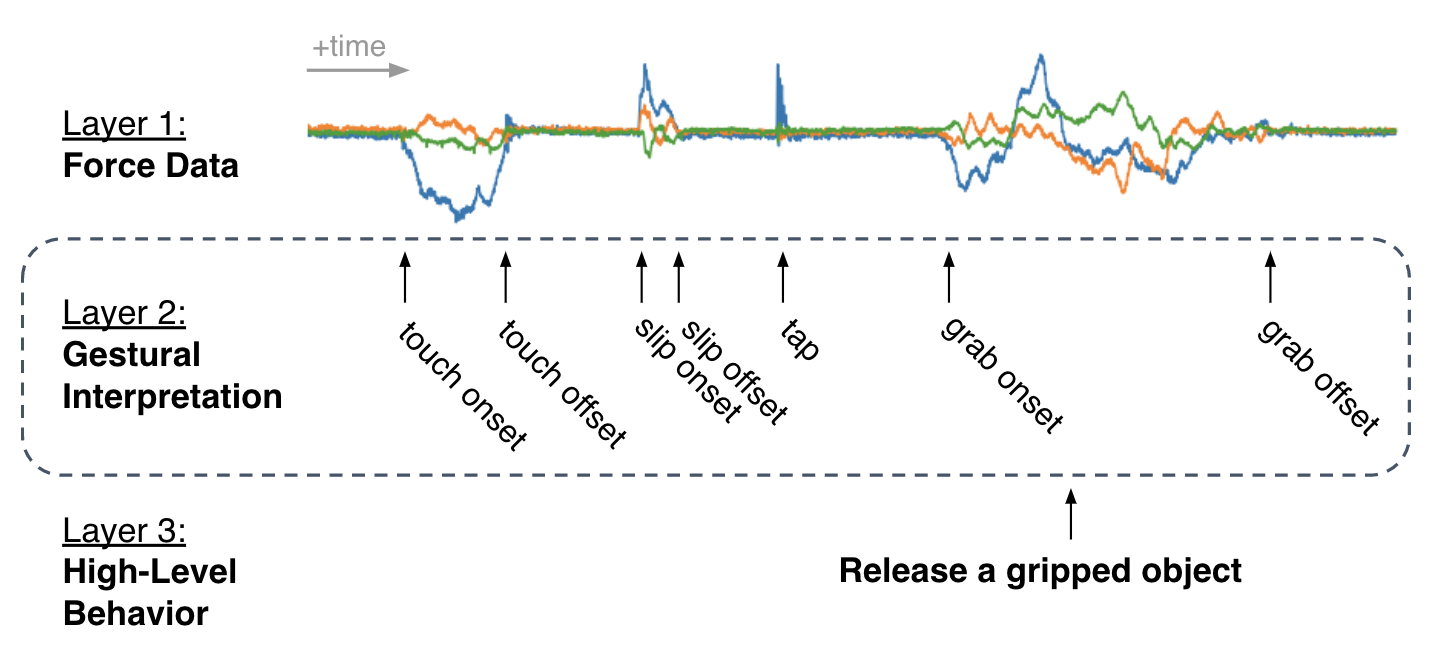}
    \caption{This work aims to build a competent gestural interpretation layer, as illustrated above in relation to low-level force-torque sensor data and to high-level behavioral system decisions.  The above exemplifies the utility of this layer for the purpose of a robot-to-human object hand-off.}
    \vspace{-0.5cm}
\end{figure}


Making robotic systems broadly useful in a range of pHRI tasks requires identifying and distinguishing among many forms of contact that could occur between human and robot.  Subtle gestural differences even within a task can affect how humans perceive other humans such as in handshaking \cite{Bailenson2007_1, Bailenson2007_2, Vigni2019} or through haptic device use \cite{Brave2001}.  Recent pHRI works reference a \enquote{touch dictionary} of human-to-robot touch gestures in an extensive user study that correlated the gestures with particular emotions \cite{yohanan2012role}.  Others pare down the 30 gestures to those most relevant to social robotics:  20 gestures in \cite{cooney2014affectionate}, 14 in \cite{jung2017automatic}, 6 in \cite{silvera2012interpretation}, and 4 in \cite{alonso2017detecting}.  Further simplification of gestural classes led other researchers to consider distinguishing all intentional from all accidental human-robot contact \cite{golz2015using, briquet2019using}.  Clearly, there is no standard for human touch gesture categories in pHRI research.

Similarly, there is no standard contact sensor:  the works represent an immense variety of force-torque sensors \cite{chen2015evaluation, Konstantinova2017, jain2013improving, lee2019making}, fingertip force sensors \cite{Nagata1998}, actuator current sensors at arm joints \cite{golz2015using, briquet2019using}, contact microphones \cite{alonso2017detecting}, pressure or touch sensor arrays \cite{jung2017automatic, silvera2012interpretation}, BioTac sensors \cite{Sundaralingam2019}, and external visual sensors \cite{li2019connecting, cooney2014affectionate, Pan2019}.  Touchscreen sensor technologies accurately distinguish intentional human contact gestures; however, an array of physical limitations prevent their use on most robotic platforms for active sensing:  effectiveness through nonconductive materials (capacitive sensors), low responsiveness (resistive), unintentional activation (optical), or perceived unpleasantness (sound acoustic wave) \cite{orphanides2017touchscreen}.  Translating across different sensor readings is a challenge requiring significant data to attain reasonable accuracies \cite{Sundaralingam2019}.  Thus, comparing system performances across differences in sensors is non-trivial and reasonably requires constancy of all other variables.  Robotic demonstrations with external visual sensors in particular, while producing convincingly natural interactions, require the crutch of external sensing, which has no biological parallel and cannot be feasibly implemented on mobile platforms or in dynamic and/or unknown environments.

Like with differences in sensors, only a few prior works compare different classification algorithms against each other \cite{alonso2017detecting}, and none the authors have found compares differences in feature sets ported into the classifiers.


\subsection{Contributions and Paper Organization}
This paper provides a comprehensive, direct comparison of system performance in identifying four touch gestures with influences from prior work, testing five forms of feature sets and three classification approaches.  The sensor data used for this endeavor comes from the internal wrist force-torque sensor of a UR5e: a common, off-the-shelf platform many pHRI research groups are likely to have and use for future studies.  Fixing only the variables of sensor type (selected to be ubiquitous) and gestures present in the dataset (collected to distinctly yet broadly cover many of the 30 gestures identified in \cite{yohanan2012role}), we demonstrate strong performance.  Our conclusions on most effective feature set and classification algorithm translate to future work as a proposed benchmark.

Section \ref{sec:problem_formulation} proposes a dictionary of gestural types, and justifies its comprehensiveness against categories used in other noted work.  The section further characterizes the dataset design and collection used for the remainder of this project.  Section \ref{sec:feature_sets} describes the unique feature sets, and Section \ref{sec:algorithms} describes the explored classification algorithms.  Section \ref{sec:results} presents the results of the trained models, using combinations of feature sets and algorithms aforementioned.  We conclude in Section \ref{sec:future_work} with suggested future directions.

\section{Problem Formulation} \label{sec:problem_formulation}
\subsection{Gesture Definitions and Vocabulary} \label{sub_sec:vocab}
We developed a dictionary of human-to-robot touch gestures to differentiate between categories of human intent or preparedness for collaborative pHRI tasks.  The original 30 gesture dictionary in \cite{yohanan2012role} and alternative 20 in \cite{cooney2014affectionate} include purely social interactions (e.g. nuzzle, massage, tickle in \cite{yohanan2012role}), specify parts of a humanoid robot to which a gesture is applied (e.g. push chest, rub head, slap cheek in \cite{cooney2014affectionate}), and include lifting the entire object/robotic system (e.g. toss, swing, shake, rock in \cite{yohanan2012role}), irrelevant for goal-driven pHRI task completion.  Other works narrowed these 20-30 gestures to smaller subsets as noted in Section \ref{sub_sec:prior_work}, with categories ranging from as many as 14 to as few as 4.  We propose a new dictionary of 4 gesture definitions, the combination of which is broad enough to cover all relevant gestures of the 30 in \cite{yohanan2012role} and 14 in \cite{jung2017automatic} and goes beyond the scope of the 6 in \cite{silvera2012interpretation} and 4 in \cite{alonso2017detecting}.  This definition set is the broadest coverage of distinct human touch gestures within a minimal number of categories we have found in the literature.

\begin{table}[t]
    \vspace{0.15cm}
    \caption{Our Gesture Definitions Compared to Others}
    \centering
    \begin{tabular}{|l|l|l|l|l|l|}
    \hline
    \textbf{Ours} & \textbf{\cite{yohanan2012role}} & \textbf{\cite{jung2017automatic}} & \textbf{\cite{silvera2012interpretation}} & \textbf{\cite{alonso2017detecting}} \\ \hline
    Tap & \makecell[tl]{hit \\ pat \\ poke \\ slap \\ tap} & \makecell[tl]{hit \\ pat \\ poke \\ slap \\ tap} & \makecell[tl]{pat \\ slap \\ tap} & \makecell[tl]{slap \\ tap} \\ \hline
    Touch & \makecell[tl]{contact \\ hold\tablefootnote{The \enquote{hold} category in \cite{yohanan2012role} can be interpreted in two ways and thus fits into two of our gesture definitions.\label{footnote:first}} \\ lift \\ press \\ push} & \makecell[tl]{press} & \makecell[tl]{push} & \\ \hline
    Grab & \makecell[tl]{cradle \\ grab \\ hold\footref{footnote:first} \\ pinch \\ pull \\ squeeze} & \makecell[tl]{grab \\ pinch \\ squeeze} & & \\ \hline
    Slip & \makecell[tl]{rub \\ scratch \\ stroke} & \makecell[tl]{rub \\ scratch \\ stroke} & \makecell[tl]{scratch \\ stroke} & \makecell[tl]{stroke} \\ \hline
    \makecell[tl]{Excluded \\ from our \\ definitions} & \makecell[tl]{finger idly \\ hug \\ kiss \\ massage \\ nuzzle \\ pick \\ rock \\ shake \\ swing \\ tickle \\ toss \\ tremble} & \makecell[tl]{massage \\ tickle} & & \makecell[tl]{tickle} \\ \hline
    \end{tabular}
    \label{table:gestures}
    \vspace{-0.5cm}
\end{table}

The four gestures outlined in our proposed dictionary are 1) tap, 2) touch, 3) grab, and 4) slip.  See Table \ref{table:gestures} for direct comparison of terminology across other sources.  This correspondence allows us to calculate comparison performances in \cite{jung2017automatic} against our categories (see Section \ref{sec:results}).

Our gesture category definitions are defined as follows:

\subsubsection{Tap Gesture}
A tap is an impulse-like event, wherein contact is made and broken in quick succession, occurring between the robot and any part of the human hand.

\subsubsection{Touch Gesture}
A touch is analogous to a tap with indefinitely longer duration.

\subsubsection{Grab Gesture}
A grab is defined as more than one touch occurring simultaneously from multiple directions.  This subtlety between touches and grabs becomes critical in human-robot handovers, in which touches on their own may not maintain control of an object.  The duration of a grab gesture, like that of a touch, can be indefinite.

\subsubsection{Slip Gesture}
A slip involves contact motion between the human and the robot.  A slip may involve contact from one primary direction (like touch) or from two directions (like grab).  The duration of a slip gesture is also indefinite.

As is clear from the existing literature, there is no perfect way to categorize human touches due to their analog nature.  Prior works highlight the necessity of defining fewer classes for the purpose of improving accuracy, emphasizing that class definitions can ultimately be a trade-off with application \cite{jung2017automatic}.  In the effort of enabling pHRI tasks, we construct our four categories to consider a wide variety of contacts, while still simplistically combining nuanced differences into functional groups.  These groups could 1) be intentionally leveraged to communicate from human to robot, 2) be combined hierarchically with other algorithms to discern intra-group differences, and 3) cover a broad scope of potential unorchestrated human-robot interactions.

\subsection{State-Based and Transition-Based Approaches} \label{sub_sec:state_vs_trans}
A robot can keep track of haptic gestures throughout time series data in two ways:  by identifying the current state (i.e. determining gesture occurrence and type) or by identifying transitions between states (i.e. recognizing a change of gesture occurrence or type).  We refer to the start and end transition of a gesture as the \enquote{onset} and \enquote{offset} of the gesture, respectively.  A system that implements a state-based touch gesture identification model can infer the transition events, as can a transition-based model infer the current state.  Thus, a system only has to implement one of these strategies accurately to effectively gain the same information.  The models developed herein take the state-based approach.

\begin{figure}[t]
    \centering
    \vspace{0.15cm}
    \includegraphics[width=0.75\linewidth]{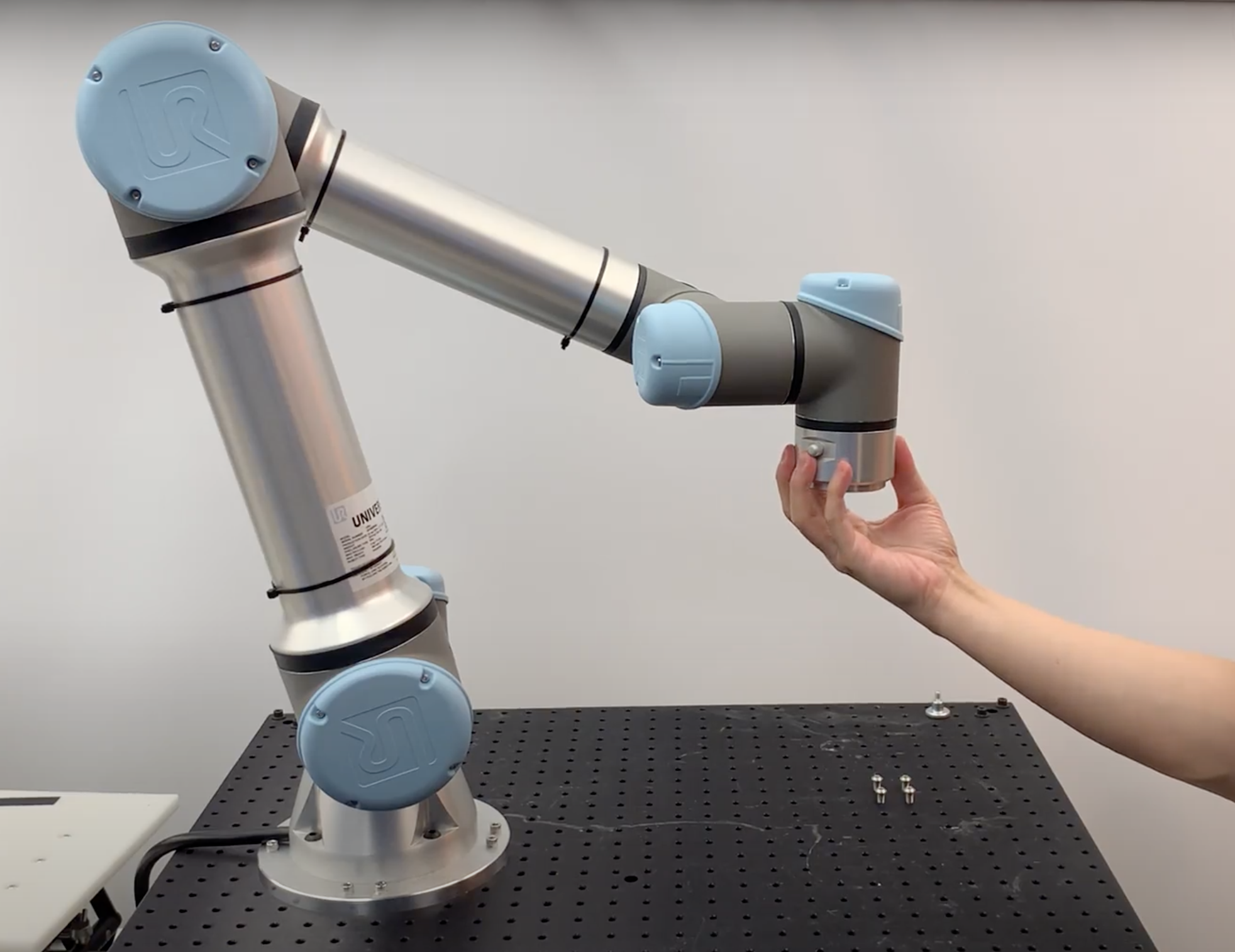}
    \vspace{-0.1cm}
    \caption{An example of a user initiating a grab gesture to the robotic arm's end effector during the data collection experiments.  Note that this visual depicts the scenario in which the end effector is removed and contact occurs at the robotic arm's distal plate.  Half of the dataset involved contact with the robot through an installed end effector.}
    \label{fig:gesture_example}
    \vspace{-0.5cm}
\end{figure}

\subsection{Dataset Generation} \label{sub_sec:dataset}
The data used in this project was collected in the Salisbury Robotics Lab, using a UR5e collaborative robotic arm.  The internal six-axis wrist force-torque sensor and its design as a collaborative system renders the UR5e an attractive option for pHRI applications in which a robot's sense of force-based touch perception could offer extensive new capabilities.

The dataset used for training and testing the models is composed of force readings from two user subjects who were instructed to apply each of the four gesture types to the end of the robotic arm.  It is our hope that this dataset will be expanded in the future to ensure further generalizability of the results.  Half of the data was collected with an end effector installed, while the other half omitted any end effector, allowing the users to touch the plate at the end of the robotic arm directly, as in Figure \ref{fig:gesture_example}.  This split is intentional, as these techniques are designed to be agnostic to the installed end effector, as long as it is rigid enough to transfer forces to the wrist with low damping and latency.  In all cases, the robotic arm was stationary in a configuration as in Figure \ref{fig:gesture_example}.  Users were instructed to vary which and what part of their hand as well as the pressure, duration, and location of the contact.

The dataset represents 24 minutes of force-torque sensor readings sampled at 500 Hz.  The raw data was filtered to remove noise, then cut into 600 millisecond duration overlapping snippets, generated at 100 millisecond increments.  The chosen duration distinguishes taps from touches (long enough for entire taps to appear within a single snippet whereas touches took longer for all of the collected examples) and is appropriately short enough for a robot employing this technique to make sufficiently real-time decisions.

The models described herein used the three force channels only, since the focus of this work is to identify the type of contact gesture occurring at the end effector.  Future iterations could incorporate the torque channels to improve accuracy in this endeavor.  Six-axis force-torque sensors provide enough information to localize single point contacts \cite{salisbury1984interpretation}, so this could be a possibility with our dataset in future studies.

\begin{figure}[b]
    \centering
    \vspace{-0.5cm}
    \includegraphics[width=0.9\linewidth]{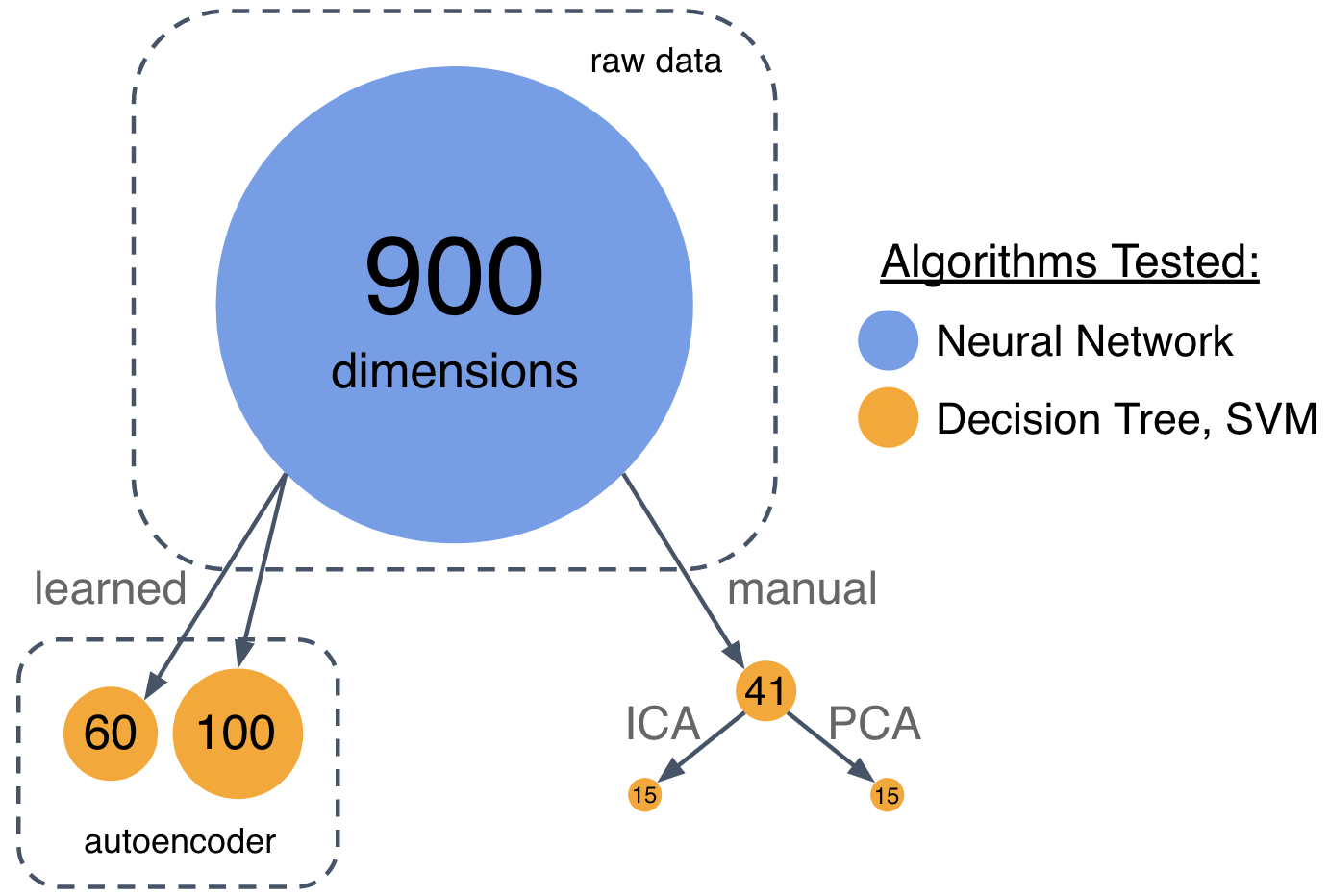} 
    \caption{Six feature sets and three classification algorithms were tested, for a total of 11 models.  The above shows the feature sets' relationships, relative dimensionality, and algorithm(s) employed on them.}
    \label{fig:feature_sets}
\end{figure}

\section{Feature Sets}  \label{sec:feature_sets}
This work explores several feature sets, including the full filtered data itself.  The other five feature sets employed by some of these models decrease the dimensionality and emulate other techniques in the literature.  The permutations of feature sets and algorithms employed in this research are illustrated in Figure \ref{fig:feature_sets}.

\subsection{Autoencoder Bottleneck} \label{sub_sec:autoencoder_featues}
Two of the five reduced-size feature sets were generated using a reconstruction convolutional autoencoder model.  These models learn a compressed version of the original snippet at the network's bottleneck \cite{hinton2006reducing}.  Manual verification determined the efficacy of these models by comparing their outputs to their inputs.  Two feature sets were constructed in this way:  one of 100 dimensions for a compression ratio of 9, and another of 60 dimensions for a compression ratio of 15, described by the size of the bottleneck layer.  We explored higher compression ratios (20+), but those models did not converge to give a reconstructable output for our dataset.

\subsection{Manual Feature Selection}  \label{sub_sec:manual_features}
\begin{figure}[t]
    \centering
    \vspace{0.1cm}
    \includegraphics[width=\linewidth]{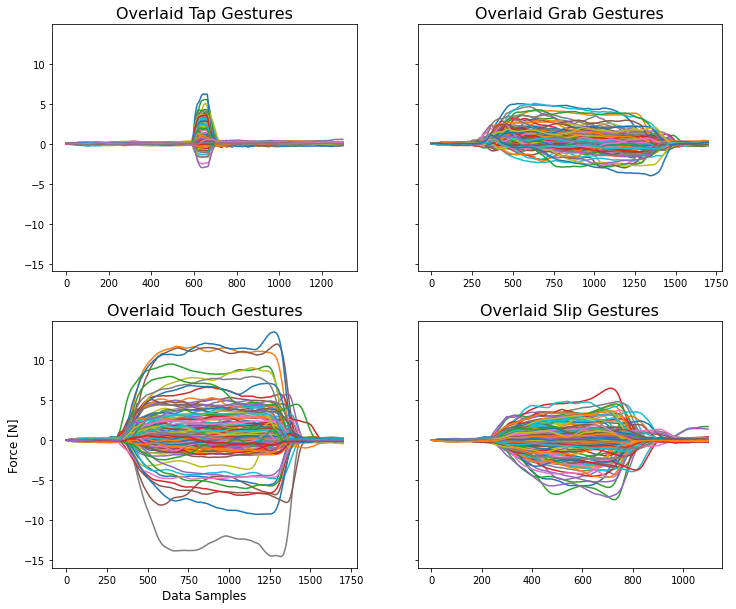}
    \caption{Groupings of each gesture type of similar duration as they appear through force sensor readings.  All x and y axis scales are shared.}
    \label{fig:overlaid_gestures}
    \vspace{-0.5cm}
\end{figure}

The remaining three reduced-size feature sets were constructed manually based on distinguishing characteristics between the gestures in the force-torque sensor readings (see Figure \ref{fig:overlaid_gestures}).  These visuals provided insight into characteristics that could both unify examples of the same gesture and discriminate between gestures.

41 features were designed to capture some of these critical characteristics.  These features correlate with other feature sets employed on similar data in prior works \cite{jung2017automatic, silvera2012interpretation, alonso2017detecting, golz2015using}.  The initial force level of every data snippet is subtracted out prior to feature calculation such that the features are invariant in the presence of a DC bias.  Prior to their usage in training the subsequent algorithms, each feature was normalized to span $[0, 1]$.  The features ($f_i$) are listed in Table \ref{table:manual_features}, which indicates which quantities store their maximum and minimum as features, and which quantities store their values per three force channels as features.

We omit further discussion of calculating the simplest features:  slope ($f_1$-$f_5$) and its duration ($f_6$-$f_8$), curvature ($f_7$-$f_{11}$) and its duration ($f_{12}$), standard deviation ($f_{13}$), and third-order polynomial fit coefficients ($f_{14}$-$f_{22}$).  The third-order polynomial fit coefficients are calculated using least-square error method; the offset is ignored while the other coefficients are stored for each force channel ($f_{14}$-$f_{22}$).  The remaining features are elaborated below.

\begin{table}[t]
    \vspace{0.15cm}
    \caption{Manual Features Computed on Force Signal}
    \centering
    \begin{tabular}{|l|p{3.6cm}|c|c|}
    \hline
    \textbf{\#} & \textbf{Feature} & \textbf{\hspace{-0.1cm}Max/Min\hspace{-0.1cm}} & \textbf{\hspace{-0.1cm}Per Channel\hspace{-0.1cm}} \\ \hline 
    1-2 & Peak Slope & \ding{51} & \\ \hline 
    3-5 & Average Slope & & \ding{51} \\ \hline 
    6-8 & High Slope Duration & & \ding{51} \\ \hline
    7-8 & Peak Curvature (2nd derivative) & \ding{51} & \\ \hline
    9-11 & Average Curvature & \ding{51} & \\ \hline
    12 & High Curvature Duration & & \\ \hline
    13 & Standard Deviation & & \\ \hline 
    14-22 & Third-Order Polynomial Fit Coefficients & & \ding{51} \\ \hline 
    23-24 & Window Main-Lobe Matching (Upward Tap Profile) & \ding{51} & \\ \hline
    25-26 & Window Main-Lobe Matching (Downward Tap Profile) & \ding{51} & \\ \hline
    27-32 & Spectrogram Slice Linear Fit & \ding{51} & \ding{51} \\ \hline
    33-35 & Spectral Rolloff & & \ding{51} \\ \hline
    36-38 & Spectral Centroid & & \ding{51} \\ \hline
    39-41 & Spectral Flatness & & \ding{51} \\ \hline
    \end{tabular}
    \label{table:manual_features}
    \vspace{-0.5cm}
\end{table}

\subsubsection{Tap Profile}
Their impulse-like nature means tap gestures all resemble a similar profile, albeit differing in direction and magnitude.  A method that compares an input signal to a pre-selected template signal is called window main-lobe matching \cite{Rao2012}.  To implement, all snippet channel data are compared to a template tap profile.  If the data points closely resemble the template, the resulting correlation error is low.  Mathematically, a normalized correlation error $\xi$ is given by:
\begin{align}
    \xi &= \frac{\epsilon}{\sum_{t=a}^b X^2(t)}, \\
    \text{where} \quad \epsilon &= \sum_{t=a}^b [X(t) - |A|W(t)]^2 \\
    \text{and} \quad A &= \frac{\sum_{t=a}^b X(t) W(t)}{\sum_{t=a}^b W^2 (t)}.
\end{align}

$\xi$ is a normalized form of the error $\epsilon$ of an input signal $X(t)$ of length $b-a$ compared to a template signal $W(t)$, selected from the template data.  $A$ is a scaling factor to ensure scale-independence, meaning a high force tap template could still be useful in detecting a tap of low force.  Per snippet, two 320-millisecond templates -- one for an upwards tap, one for a downwards tap -- are compared against each of the force channels.  Both the lowest and highest error $\xi$ to each template are stored, yielding features $f_{23}$ through $f_{26}$.

\begin{figure}[t]
    \centering
    \vspace{0.1cm}
    \includegraphics[width=0.8\linewidth]{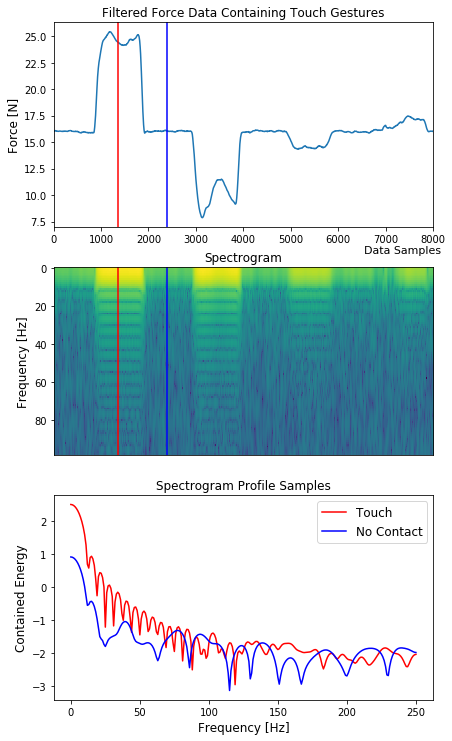}
    \vspace{-0.25cm}
    \caption{Force data (top) produces a spectrogram (center) that displays more energetic frequency content at low frequencies during touch events than during periods of no human contact.  This is captured by a linear approximation for a time slice of the frequency content.  Two example slice profiles are illustrated (bottom), with their corresponding time stamps marked in the upper two plots.}
    \label{fig:spec}
    \vspace{-0.5cm}
\end{figure}

\subsubsection{Spectrogram Slice Linear Approximation}
Noting the time series similarity between force and audio data, we employed many techniques commonly used for audio signal analysis \cite{peeters2004large}.  One such technique is short-time Fourier transform, which quantifies the frequency content throughout a time series signal.  Previous studies indicate that humans have particular tremor frequencies \cite{Stiles1967} that could be injected into a robotic system's force-torque sensor readings at the occurrence of human contact.  Figure \ref{fig:spec} demonstrates the ability of this tool to distinguish contact from non-contact, the former of which exhibits higher energies of low frequencies.

For each of the three channels and through time, a line with least squared error is computed for the spectrogram slice profile.  Of these, the maximum, minimum, and average value for the slope and offset are stored as features $f_{27}$ through $f_{32}$.

\subsubsection{Spectral Rolloff, Centroid, and Flatness}
Among several other ways of characterizing frequency content from a generated spectrogram \cite{librosa} are:
\begin{itemize}
    \item Spectral rolloff: the frequency beyond which a diminishing proportion of the total energy is contained.
    \item Spectral centroid: a measure akin to the \enquote{center of mass} of the frequency content.
    \item Spectral flatness: a quantification of tone-like versus noise-like characteristics.
\end{itemize}

The remaining nine features, $f_{33}$ through $f_{41}$, are the average values of spectral rolloff, spectral centroid, and spectral flatness across each of the three force data channels.

\subsection{Principal and Independent Component Analysis} \label{sub_sec:pca_ica}
Lower dimensional representations of the manually selected features $f_1$ through $f_{41}$ are also evaluated, via principal component analysis (PCA), as used in \cite{jain2013improving}, and independent component analysis (ICA).  Using PCA, 15 principal components are selected to maintain 95\% of the training data variance at a reduced dimensionality.  To match the PCA feature set size, ICA generated 15 independent components.

\section{Algorithms} \label{sec:algorithms}

This work directly compares several commonly used albeit infrequently justified classification algorithms.  Fixing all other factors, this direct comparison should be informative in showing the strengths and weaknesses of each approach.

\subsection{Classifiers} \label{sub_sec:classifiers}
This research evaluates three classifier algorithm types:  decision trees, support vector machines (SVM), and neural network classifiers.  While all three generally improve with larger datasets, the size of a sufficiently large dataset for neural network classifiers is particularly difficult to determine \cite{Barbedo2018} and was not guaranteed to be met by our dataset.

\begin{figure}[t]
    \centering
    \vspace{0.1cm}
    \includegraphics[width=0.8\linewidth]{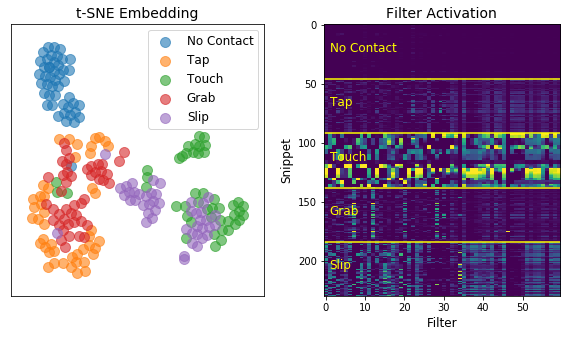}
    \vspace{-0.25cm}
    \caption{A t-SNE plot (left) of a convolutional layer for the validation set indicates clustering of gesture types, and filter activations (right) throughout the CNN demonstrates similar activation patterns within like gesture types.}
    \label{fig:tsne_and_filters}
    \vspace{-0.5cm}
\end{figure}

\begin{figure*}[t]
    \centering
    \vspace{0.1cm}
    \includegraphics[width=0.9\textwidth]{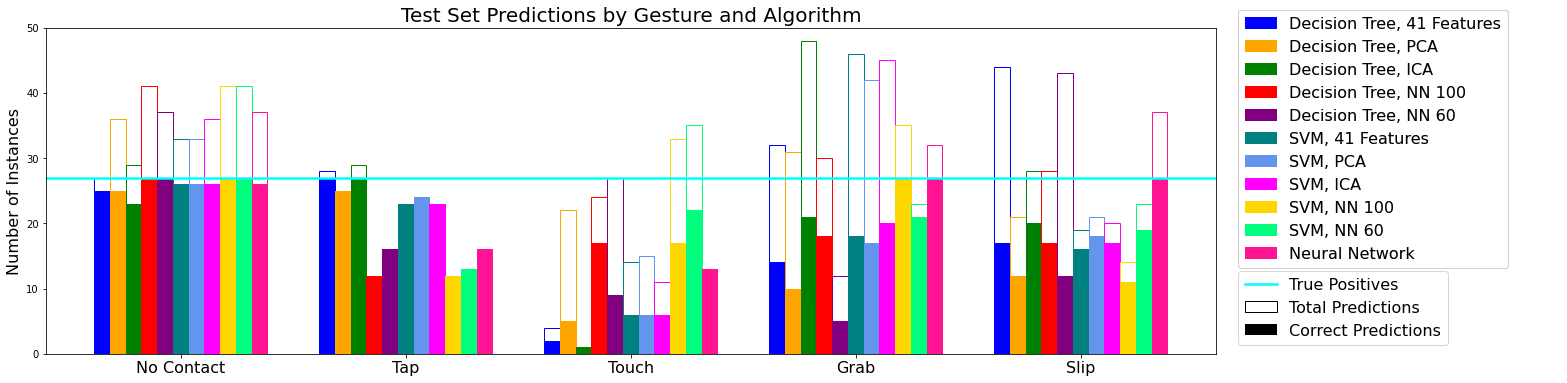}
    \vspace{-0.25cm}
    \caption{This plot shows the true and false positives for each of the 11 trained classification models for each of the 5 state-based categories on the test set.}
    \label{fig:performance_breakdown}
    \vspace{-0.2cm}
\end{figure*}

We constructed and searched over 1D CNNs where filters are learned to perform temporal convolutions.  We opted for simple linear convolutions instead of dilated convolutions \cite{lee2019making}, recurrent neural networks, or variants including LSTMs \cite{hochreiter1997long}, as our force signal application has near-sighted contexts and does not require solving long-term dependencies.  Figure \ref{fig:tsne_and_filters} shows validation set t-SNE and filter activation plots.

\begin{figure}[t]
    \centering
    \includegraphics[width=0.9\linewidth]{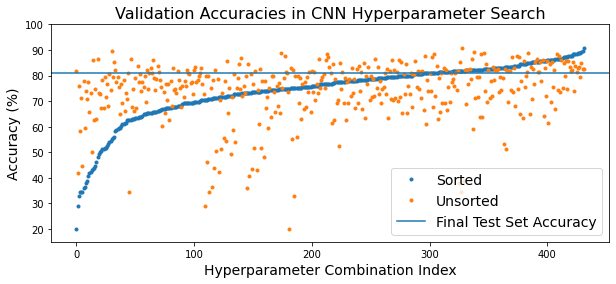}
    \caption{Plot of highest validation accuracies obtained at each index through the initial hyperparameter search round.  The sorted points display the same data in ascending order as the unsorted points.  The test set accuracy line indicates above it all of the models that overfit to the training data.}
    \label{fig:training}
    \vspace{-0.5cm}
\end{figure}

A neural architecture search \cite{elsken2018neural} was performed through two selection rounds: the first round stored several models with the highest validation accuracy over 200 epochs, and a second round trained that limited set of models for an additional 1000 epochs at a 10x slower learning rate, saving the lowest validation loss thereafter.  The hyperparameter search (over network topology, learning rate, filters, and dropout rates) was necessary and non-trivial, as indicated by the sporadic and frequently overfitting results pictured in Figure \ref{fig:training}.  All optimization used the Adam optimizer \cite{adam}.  The final model has 16 filters, an initial learning rate of 1e-4, and 5 convolution layers, each followed by a dropout layer to mitigate overfitting \cite{srivastava2014}, with dropout rates of 0.5 after the first 3 convolution layers and 0.1 after the last 2.

Hyperparameter searches were performed for each feature set on the decision tree and SVM algorithms.  The hyperparameters that yielded the highest validation set accuracy at the end of the second training round were selected.  Those hyperparameters are:  \textit{Decision Tree:} max. tree depth, min. samples per leaf, min. samples per split, random state variable (for repeatability);  \textit{SVM:} regularization parameter, kernel type, kernel coefficient (for 'rbf', 'poly', and 'sigmoid' kernels), decision shape, random state variable.

\subsection{Models for Autoencoder Feature Sets} \label{sub_sec:autoencoder_models}
The autoencoders trained to generate the two learned feature sets discussed in Section \ref{sub_sec:autoencoder_featues} are convolutional neural networks (CNNs) \cite{kiranyaz20191d, ronneberger2015u} built using Keras \cite{keras}.  Both autoencoder models use 128 filters, optimized with the Adam optimizer \cite{adam} with a learning rate of 1e-4 over 1000 epochs with a batch size of 128, saving the weights corresponding to the lowest validation loss.

\begin{table}[b]
    \vspace{-0.2cm}
    \caption{Test Set Accuracies by Feature Set and Algorithm}
    \centering
    \begin{tabular}{|l|c|c|c|}
    \hline
    \textbf{Feature Set} & \textbf{Decision Tree} & \textbf{SVM} & \textbf{NN}  \\ \hline
    \textbf{Manual}          & 63\% & 66\% & - \\ \hline
    \textbf{PCA}             & 57\% & 67\% & - \\ \hline
    \textbf{ICA}             & 68\% & 68\% & - \\ \hline
    \textbf{Autoencoder 60}  & 67\% & 70\% & - \\ \hline
    \textbf{Autoencoder 100} & 51\% & 76\% & - \\ \hline
    \textbf{Force Data}      & -    & -    & 81\% \\ \hline
    \end{tabular}
    \label{table:performance_accuracy}
\end{table}

\section{Results} \label{sec:results}
The training set contains 319 snippets for each of the five states, totaling 1,595 examples.  A held-out validation set contains 46 snippets for each category, totaling 230 examples.  Performance on this validation set determined the hyperparameter values employed per algorithm and feature set combination.  The separate test set contains 27 snippets per category, for 135 examples.

\begin{table}[]
    \caption{Confusion Matrices}
    \centering
    \begin{tabular}{| c | l | c | c | c | c | c |}
    \hhline{~~-----}
    \multicolumn{2}{c|}{\multirow{2}{*}{}} & \multicolumn{5}{c|}{\textbf{Predicted Gesture}} \\ \hhline{|~~|-----}
    \multicolumn{2}{c|}{} & \textbf{a} & \textbf{b} & \textbf{c} & \textbf{d} & \textbf{e}  \\ \hhline{--=====}
    \multicolumn{7}{|c|}{\textbf{Decision Tree with ICA}} \\ \hline
    \parbox[t]{2mm}{\multirow{5}{*}{\rotatebox[origin=c]{90}{\textbf{Actual Gesture}}}} & \textbf{No Contact (a)} & \cellcolor{green!46}23 & 0 & 0 & \cellcolor{red!6}3 & \cellcolor{red!2}1 \\ \hhline{|~|------}
    & \textbf{Tap (b)} & 0 & \cellcolor{green!54}27 & 0 & 0 & 0 \\ \hhline{|~|------}
    & \textbf{Touch (c)} & \cellcolor{red!8}4 & 0 & \cellcolor{green!2}1 & \cellcolor{red!38}19 & \cellcolor{red!6}3 \\ \hhline{|~|------}
    & \textbf{Grab (d)} & \cellcolor{red!4}2 & 0 & 0 & \cellcolor{green!42}21 & \cellcolor{red!8}4 \\ \hhline{|~|------}
    & \textbf{Slip (e)} & 0 & \cellcolor{red!4}2 & 0 & \cellcolor{red!10}5 & \cellcolor{green!40}20 \\ \hline \hline
    \multicolumn{7}{|c|}{\textbf{SVM with NN 100}} \\ \hline
    \parbox[t]{2mm}{\multirow{5}{*}{\rotatebox[origin=c]{90}{\textbf{Actual Gesture}}}} & \textbf{No Contact (a)} & \cellcolor{green!54}27 & 0 & 0 & 0 & 0 \\ \hhline{|~|------}
    & \textbf{Tap (b)} & \cellcolor{red!28}14 & \cellcolor{green!24}12 & 0 & \cellcolor{red!2}1 & 0 \\ \hhline{|~|------}
    & \textbf{Touch (c)} & 0 & 0 & \cellcolor{green!34}17 & \cellcolor{red!14}7 & \cellcolor{red!6}3 \\ \hhline{|~|------}
    & \textbf{Grab (d)} & 0 & 0 & 0 & \cellcolor{green!54}27 & 0 \\ \hhline{|~|------}
    & \textbf{Slip (e)} & 0 & 0 & \cellcolor{red!32}16 & 0 & \cellcolor{green!22}11 \\ \hline \hline
    \multicolumn{7}{|c|}{\textbf{Neural Network}} \\ \hline
    \parbox[t]{2mm}{\multirow{5}{*}{\rotatebox[origin=c]{90}{\textbf{Actual Gesture}}}} & \textbf{No Contact (a)} & \cellcolor{green!52}26 & 0 & 0 & \cellcolor{red!2}1 & 0 \\ \hhline{|~|------}
    & \textbf{Tap (b)} & \cellcolor{red!22}11 & \cellcolor{green!32}16 & 0 & 0 & 0 \\ \hhline{|~|------}
    & \textbf{Touch (c)} & 0 & 0 & \cellcolor{green!26}13 & \cellcolor{red!8}4 & \cellcolor{red!20}10 \\ \hhline{|~|------}
    & \textbf{Grab (d)} & 0 & 0 & 0 & \cellcolor{green!54}27 & 0 \\ \hhline{|~|------}
    & \textbf{Slip (e)} & 0 & 0 & 0 & 0 & \cellcolor{green!54}27 \\ \hline
    
    \end{tabular}
    \label{table:confusion}
    \vspace{-0.5cm}
\end{table}

The held-out test set accuracy results are given in Table \ref{table:performance_accuracy}, and a breakdown of true and false positives across all models and gesture types is plotted in Figure \ref{fig:performance_breakdown} (note that the cyan \enquote{True Positives} horizontal line applies to all gestures, given the balanced set).  Despite a small dataset, the best performing algorithm is the neural network classifier, which uses the force data directly as its feature set, obtaining a test set accuracy of 81\%.

Using reported data from \cite{jung2017automatic} of predictions among their 14 classes, we are able to compare our results to this prior work, which used an SVM algorithm on a manually-constructed feature set of 54 dimensions.  Combining their categories into our definitions as outlined in Table \ref{table:gestures} and correcting for the resulting unbalanced dataset, Jung et al. obtain a test set accuracy of 81.5\% within our gestural dictionary.\footnote{81.5\% accuracy is among the 4 categories of tap, touch, grab, and slip, excluding the no contact category due to lack of those results in \cite{jung2017automatic}.}  Notably, our algorithms matched their performance on our categories using a dataset less than a quarter of the size of theirs (1,960 examples versus their 7,805).  This similar performance to our neural network's 81\% indicates 1) the promise of force-torque sensors over more complicated sensors (such as pressure sensor arrays in \cite{jung2017automatic}) to discern our proposed categories, 2) the validity and data efficiency of our four gesture definitions against those of other works, 3) the nonessential role of feature computation over using the raw data, and 4) the reproducibility of our results.

After the neural network with the raw data, the next best performance is 76\% with the SVM algorithm with the autoencoder bottleneck feature set of dimension 100.  These top neural network and SVM models perform markedly better than the best decision tree model, which obtains 68\% accuracy using the ICA feature set.  Thus, the transparency of a decision tree algorithm comes at the cost of accuracy.

Confusion matrices for the top  performing algorithms of each category (decision tree, SVM, neural network) are presented in Table \ref{table:confusion}.  These matrices indicate the weaknesses of each model.  For example, the highest performing decision tree model's accuracy is skewed upward because the confusion matrix (top in Table \ref{table:confusion}) indicates this particular model seldom predicts touch gestures, inflating the likelihood of correctly guessing the other four categories.  The model misclassified most of the actual touch gestures as grabs.

The neural network and SVM models displayed similar patterns of misidentification.  For the neural network, approximately half of all touch gestures were misclassified, either as slip or grab gestures.  Additionally, while still correctly classifying the majority of tap gestures, the neural network model incorrectly labeled a significant portion as no contact.  Tap gestures had some of the highest recall in the training and validation sets, so this difference in the test set was unexpected.  Further inspection indicated that several of the tap instances in the test set were particularly low in amplitude in comparison to the training examples.

\section{Future work}
\label{sec:future_work}

\subsection{Suggestions for the Field}
Given our comprehensive comparison of feature sets and classification algorithms on a single dataset with a common, convenient sensor, we intend for our results to inform future developments in human touch gesture recognition.  Our ability to compare all tested combinations against each other yields the following list of conclusions otherwise absent from the previous works in the field:
\begin{enumerate}
    \item A well-tuned neural network using raw sensor data as its feature set can outperform other algorithms and feature sets, even using a medium-sized dataset (on the order of 1000 examples).  Design and computational effort in calculating manual features can be unnecessary.
    \item With a modest number of classes, complex, fragile tactile sensors can be met and even surpassed in performance by simple force-torque sensors, which are conveniently built into most pHRI robotic platforms.
    \item Our gesture dictionary outlines four broad yet distinct categories, covering as much scope in fewer categories than several previously proposed definition sets.  The accuracies obtained in our work are reproducible since they match those reported by others, after correcting for categorization labels.
\end{enumerate}

\subsection{Increase Accuracy and Robustness}
Using our conclusions to make progress in pHRI demonstrations, we intend to pursue further improvements in our classification methods.  Techniques such as distillation \cite{hinton2015distilling}, majority vote across multiple algorithm outputs, and data augmentation can improve performance using the existing dataset.  These models can be made more robust by expanding the dataset \cite{elsken2018neural, ellis1999size} through more human experiments.  A larger dataset in concert with distillation and/or data augmentation will likely improve the accuracy and generalizability of these developed models.  Sensor fusion is another attractive option to expand the capabilities of this approach \cite{li2019connecting, lee2019making}.



\subsection{Active Exploration of Environment}
The developed models so far identify contact events inflicted by a human on a robot.  There exists enormous opportunity for the robot to learn more about its environment and human collaborators if it performs its own active exploration.  For instance, a robotic arm could determine a measure of stability of a human's grasp by actively quantifying the mechanical impedance the human imposes through the grasped object.  Such a measure could be critical in human-robot handovers of fragile objects.

\section{Conclusion}
With results that match previous research, this work provides enlightenment on how to design feature sets and classification algorithms for the purpose of interpreting human contact.  Common and inexpensive haptic sensors can obtain high accuracies, comparable to the performance of complex tactile sensor arrays.  Equipping a robotic system with the ability to understand aspects of human touch can enable them to collaborate effectively and efficiently, one day with as much elegance as physical human-human interactions.

\section*{Acknowledgments}
This work was supported by the National Defense Science and Engineering Graduate Fellowship and in part by the Amazon Research Awards Program.

\bibliographystyle{IEEEtran}
\bibliography{refs.bib}

\end{document}